\pdfoutput=1

\documentclass[11pt]{article}

\usepackage[final]{acl}
\usepackage{enumitem}
\usepackage{times}
\usepackage{latexsym}

\usepackage[T1]{fontenc}

\usepackage[utf8]{inputenc}

\usepackage{microtype}

\usepackage{inconsolata}

\usepackage{graphicx}
\usepackage{booktabs}
\usepackage{multirow}

\usepackage{amsmath}
\usepackage{amssymb}

\usepackage{fvextra}
\usepackage{pifont}
\usepackage{footmisc}
\definecolor{hint}{RGB}{237, 175, 31}
\definecolor{error}{RGB}{255, 0, 0}

\title{CrystalICL: Enabling In-Context Learning for Crystal Generation
}

\author{Ruobing Wang \\
  Jilin University \\
  \texttt{wangrb25@mails.jlu.edu.cn} \\\And
  Qiaoyu Tan \\
  New York University Shanghai \\
  \texttt{qiaoyu.tan@nyu.edu} \\\And
  Yili Wang \\
  Jilin University \\
  \texttt{yiliwang@mails.jlu.edu.cn} \\\AND
  Ying Wang \\
  Jilin University \\
  \texttt{wangying2010@jlu.edu.cn} \\\And
  Xin Wang\textsuperscript{\ding{66}} \\
  Jilin University \\
  \texttt{xinwang@jlu.edu.cn} \\}

\begin{document}
\maketitle
\newcommand\blfootnote[1]{%
\begingroup
\renewcommand\thefootnote{}\footnote{#1}%
\addtocounter{footnote}{-1}%
\endgroup
}

\begin{abstract}
Designing crystal materials with desired physicochemical properties remains a fundamental challenge in materials science. While large language models (LLMs) have demonstrated strong in-context learning (ICL) capabilities, existing LLM-based crystal generation approaches are limited to zero-shot scenarios and are unable to benefit from few-shot scenarios. In contrast, human experts typically design new materials by modifying relevant known structures which aligns closely with the few-shot ICL paradigm. Motivated by this, we propose CrystalICL, a novel model designed for few-shot crystal generation. Specifically, we introduce a space-group based crystal tokenization method, which effectively reduces the complexity of modeling crystal symmetry in LLMs. We further introduce a condition-structure aware hybrid instruction tuning framework and a multi-task instruction tuning strategy, enabling the model to better exploit ICL by capturing structure–property relationships from limited data. Extensive experiments on four crystal generation benchmarks demonstrate the superiority of CrystalICL over the leading baseline methods on conditional and unconditional generation tasks.
\blfootnote{\textsuperscript{\ding{66}} Corresponding author}
\end{abstract}

\section{Introduction}

The design and discovery of crystal materials with specific physicochemical properties have remained a long-standing issue in the field of materials design. The development of novel crystal materials plays a crucial role in advancing technologies such as batteries, semiconductors, and catalysis \cite{butler2018machine, desiraju2002cryptic}. While traditional methods based on Density Functional Theory (DFT) \cite{kohn1965self} have proven effective, they are often computationally expensive and time-consuming. In contrast, deep learning techniques \cite{CDVAE, DiffCSP, DiffCSPPP, CrystalLM, CrystalLLM} have emerged as a powerful alternative, enabling the efficient prediction of potentially stable crystal material structures and significantly accelerating the discovery process.

\begin{figure}[t!]
    \centering
    \includegraphics[width=\columnwidth]{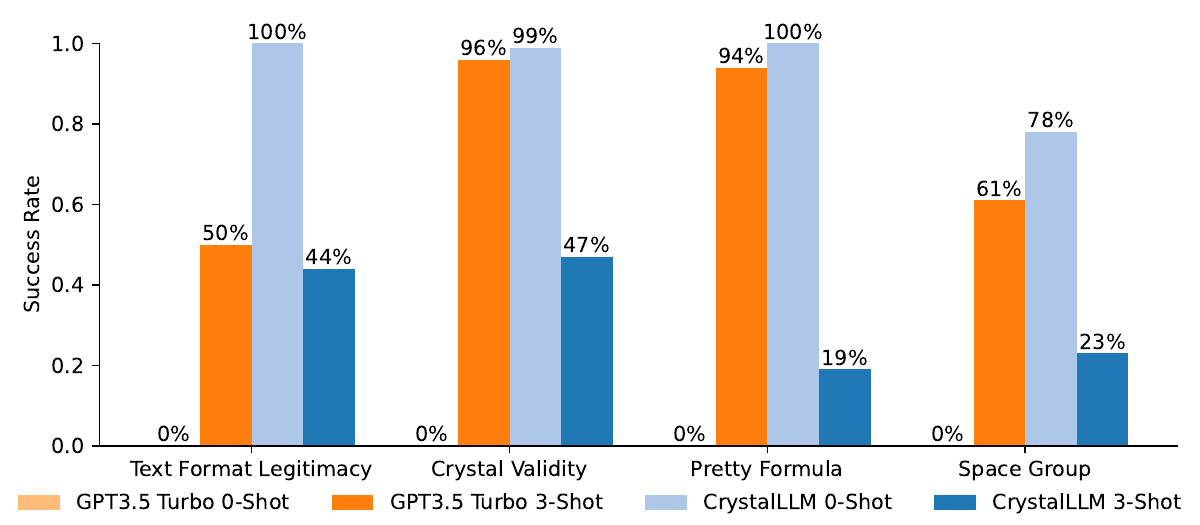}
    \vspace{-2.0em}
    \caption{The conditional crystal generation performance of GPT-3.5 Turbo and CrystalLLM on P5.} \label{ExpMov}
    \vspace{-2.0em}
\end{figure}

In recent years, the successful adaptation of LLMs in drug discovery \cite{drugdis} and protein structure prediction \cite{lin2023evolutionary} has inspired growing interest in leveraging pretrained LLMs for crystal generation tasks. 
Among these efforts, CrystalLLM \cite{CrystalLLM}, which is fine-tuned on Llama-2 \cite{llama2}, has demonstrated competitive performance in crystal generation. However, it does not fully inherit the in-context learning (ICL) capabilities of LLMs. 
These capabilities are essential for emulating the expert-driven workflow in material discovery.

Specifically, human experts typically begin with a small set of known materials that share similar properties and modify their composition or structure to achieve new design objectives. This process closely mirrors the few-shot in-context learning paradigm, where models generate new structures by referencing a limited number of relevant examples. To verify this limitation, we compare GPT-3.5 Turbo and CrystalLLM on the P5 \cite{P5} dataset under 0-shot and 3-shot prompts. As shown in Fig.~\ref{ExpMov}, GPT-3.5 struggles in the 0-shot setting but improves markedly with 3-shot prompts, demonstrating strong ICL behavior. In contrast, CrystalLLM performs worse in the 3-shot setting than in 0-shot setting, indicating its limited ability to benefit from in-context examples.

Motivated by these limitations, we aim to bridge the gap in applying ICL to crystal generation by proposing CrystalICL—the first crystal generation model explicitly designed to inherit and leverage the ICL-driven generalization capabilities of LLMs.
To this end, we first introduce \textbf{Space-group based Crystal Tokenization} (SGS), a novel method that transforms 3D crystal structures into 1D text. Compared with traditional XYZ-format crystal structure text~\cite{LMCH}, SGS significantly improves LLMs' ability to capture and model crystal symmetry. Next, we present the \textbf{Condition-Structure Aware Hybrid Crystal Instruction Tuning} framework, which incorporates three selection strategies to identify the most relevant crystal examples for downstream tasks. This framework effectively improves the model’s few-shot generation capability by allowing it to take advantage of informative contextual examples. 
Finally, to explicitly guide the model in learning the correspondence between crystal structures and their properties, we introduce the \textbf{Multi-Task Crystal Instruction Tuning} strategy. This approach integrates crystal property prediction instructions into the fine-tuning process, further enhancing the model’s ability to capture structure-property relationships and improving performance in crystal generation tasks. 
Our main \textbf{contributions} are summarized as follows:

\begin{itemize}[noitemsep,leftmargin=*,label=$\star$]
\item We explore the underutilized in-context learning (ICL) capability of LLMs for crystal generation and introduce CrystalICL, the first approach to leverage LLMs' few-shot reasoning abilities for material design, enabling efficient and adaptable crystal generation. 
\item To achieve this, we propose a tailored crystal tokenization strategy and structure-aware instruction tuning mechanisms, incorporating template-based and multitask learning to collectively enhance CrystalICL's ICL reasoning capabilities.  
\item Experimental results on four publicly available datasets across diverse domains and scales demonstrate the effectiveness of CrystalICL in both zero-shot and few-shot learning scenarios.
\end{itemize}

\section{Preliminary}
\subsection{Crystal Generation}

\begin{figure*}[!ht]
    \centering
    \includegraphics[width=\textwidth]{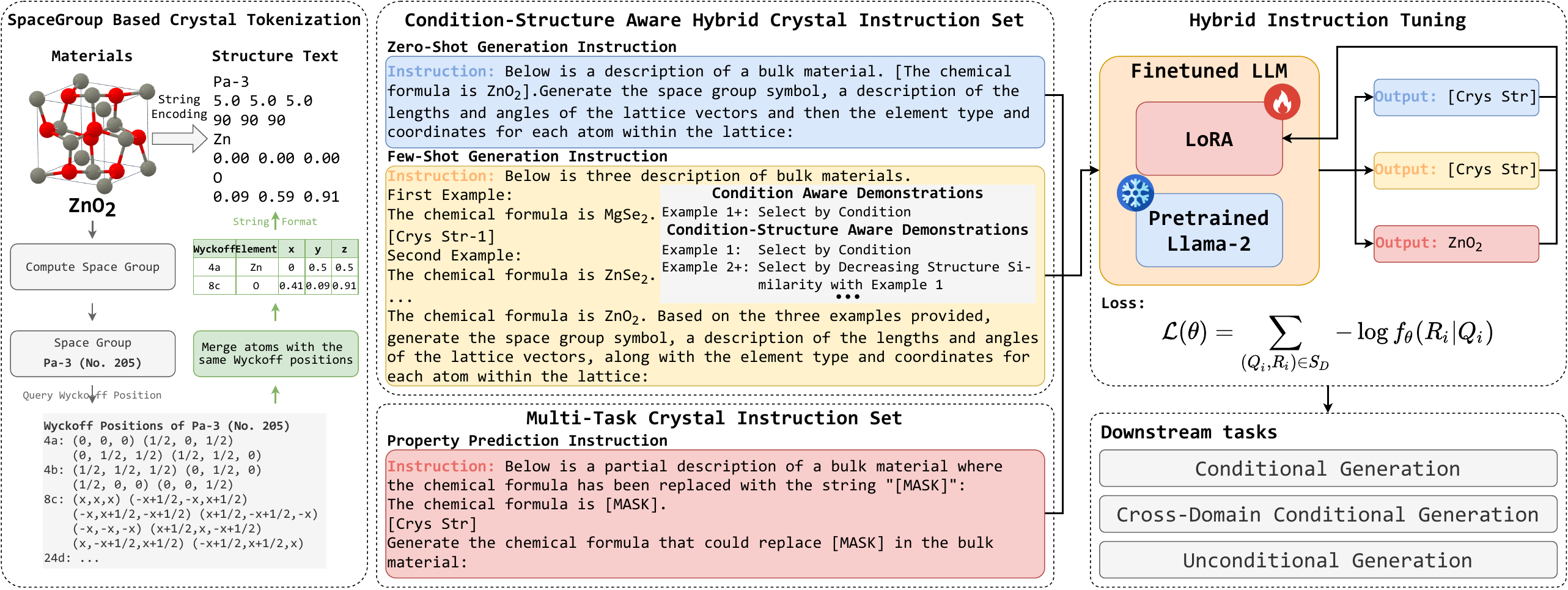}
    \vspace{-1.5em}
    \caption{The illustration of our proposed CrystalICL. CrystalICL begins by using a space-group based crystal tokenization method to transform 3D crystal structures into a text format suitable for input into LLMs. Secondly, CrystalICL constructs a condition-structure aware hybrid crystal instruction set, which includes both zero-shot and few-shot instructions. The few-shot instructions combine various example selection strategies, which query the K most relevant demonstrations for prompt design tailored to crystal generation tasks. Finally, CrystalICL incorporates a crystal property prediction instruction set, which combines with the crystal generation instruction set to form a multi-task crystal instruction set.
    } \label{Framework}
    \vspace{-1.5em}
\end{figure*}

The crystal structure is characterized by the geometry of the arrangement of particles within the unit cells. A unit cell is defined as the smallest repeating unit that preserves the full symmetry of the crystal structure. Given a unit cell containing $N$ atoms, it can be described by the triplet $\mathcal{M} = (\mathbf{A}, \mathbf{X}, \mathbf{L})$, where $\mathbf{A} = [\boldsymbol{a}_1, \boldsymbol{a}_2, \dots, \boldsymbol{a}_N]^\text{T} \in \mathbb{R}^{N \times K}$ represents a list of atomic types in one-hot encoding format ($K$ is the number of possible atomic types), $\mathbf{X} = [\boldsymbol{x}_1, \boldsymbol{x}_2, \dots, \boldsymbol{x}_N]^\text{T} \in \mathbb{R}^{N \times 3}$ contains the Cartesian coordinates of the atoms, and $\mathbf{L} = [\boldsymbol{l}_1, \boldsymbol{l}_2, \boldsymbol{l}_3]^\text{T} \in \mathbb{R}^{3 \times 3}$ is the lattice matrix that describes the periodicity of the crystal. The infinite periodic crystal structure is represented as follows:
\begin{equation}
\{(\boldsymbol{a}'_i, \boldsymbol{x}'_i) | \boldsymbol{a}'_i = \boldsymbol{a}_i, \boldsymbol{x}'_i = \boldsymbol{x}_i + \boldsymbol{k}\mathbf{L}, \forall \boldsymbol{k} \in \mathbb{Z}^{1\times3}\},
\end{equation}
where the elements of the integer vector $\boldsymbol{k}$ represent integral 3D translations along their corresponding lattice directions in $\mathbf{L}$.

In order to reflect the periodicity of the crystal structure, it is convenient to use the lattice vectors $(\boldsymbol{l}_1, \boldsymbol{l}_2, \boldsymbol{l}_3)$ to replace the standard orthogonal Cartesian basis. In this case, the Cartesian coordinates $\boldsymbol{x}=\sum^3_{i=1} f_i \boldsymbol{l}_i$ can be replaced by the fractional coordinate vector $\boldsymbol{f}=[f_1, f_2, f_3] \in [0,1)^3$. In this work, we adopt the fractional coordinate system and describe the crystal as $\mathcal{M}=(\mathbf{A},\mathbf{F},\mathbf{L})$, where the matrix $\mathbf{F}\in [0,1)^{N\times3}$ contains the fractional coordinates of all atoms in the unit cell. This work focuses  on two primary tasks:

\textbf{Conditional Crystal Generation}: Given a dataset $\{(\mathcal{M}_j, s_j)\}^n_{j=1}$, where $s_j$ denotes a specific property of $\mathcal{M}_j$, our goal is to develop a conditional generative model $p_\theta(\cdot|s)$ that generates 3D crystal structures with the specified property $s$.

\textbf{Unconditional Crystal Generation}: Given a dataset $\{\mathcal{M}_j\}^n_{j=1}$, our goal is to develop an unconditional generative model $p_\theta(\cdot)$ that can generate a collection  of crystals with a distribution similar to the training set.

\subsection{Crystal Instruction Tuning} \label{Notation}
Given a set of $n$ crystal materials $D=\{(\mathcal{M}_j, s_j)\}^n_{j=1}$, the goal of crystal instruction tuning is to fine-tune the LLM $f_\theta$ by fitting the training instruction set $S_D$ constructed from $D$ as a collection of $(\text{input},\text{output})$ pairs. The fine-tuned model is then expected to generate crystals with specific properties $s$ (conditional generation) and generate new crystal structures similar to those in the training set (unconditional generation).

\section{Method}
In this section, we introduce the proposed CrystalICL, as illustrated in Fig. \ref{Framework}.
First, we discuss a space-group based crystal tokenization method designed to reduce the complexity of modeling crystal symmetry within LLMs (in Sec. \ref{SGS}). Next, we elaborate on a condition-structure aware hybrid crystal instruction tuning framework, which effectively enhances the ICL capabilities of LLMs for crystal generation tasks (in Sec. \ref{HIns}). Finally, we introduced a multi-task crystal instruction tuning strategy, which strengthens the model’s ability to capture the relationship between crystal structures and their properties (in Sec. \ref{Inf}).

\subsection{Space-group based Crystal Tokenization}  \label{SGS}
\begin{figure}[t]
    \centering
    \includegraphics[width=\columnwidth]{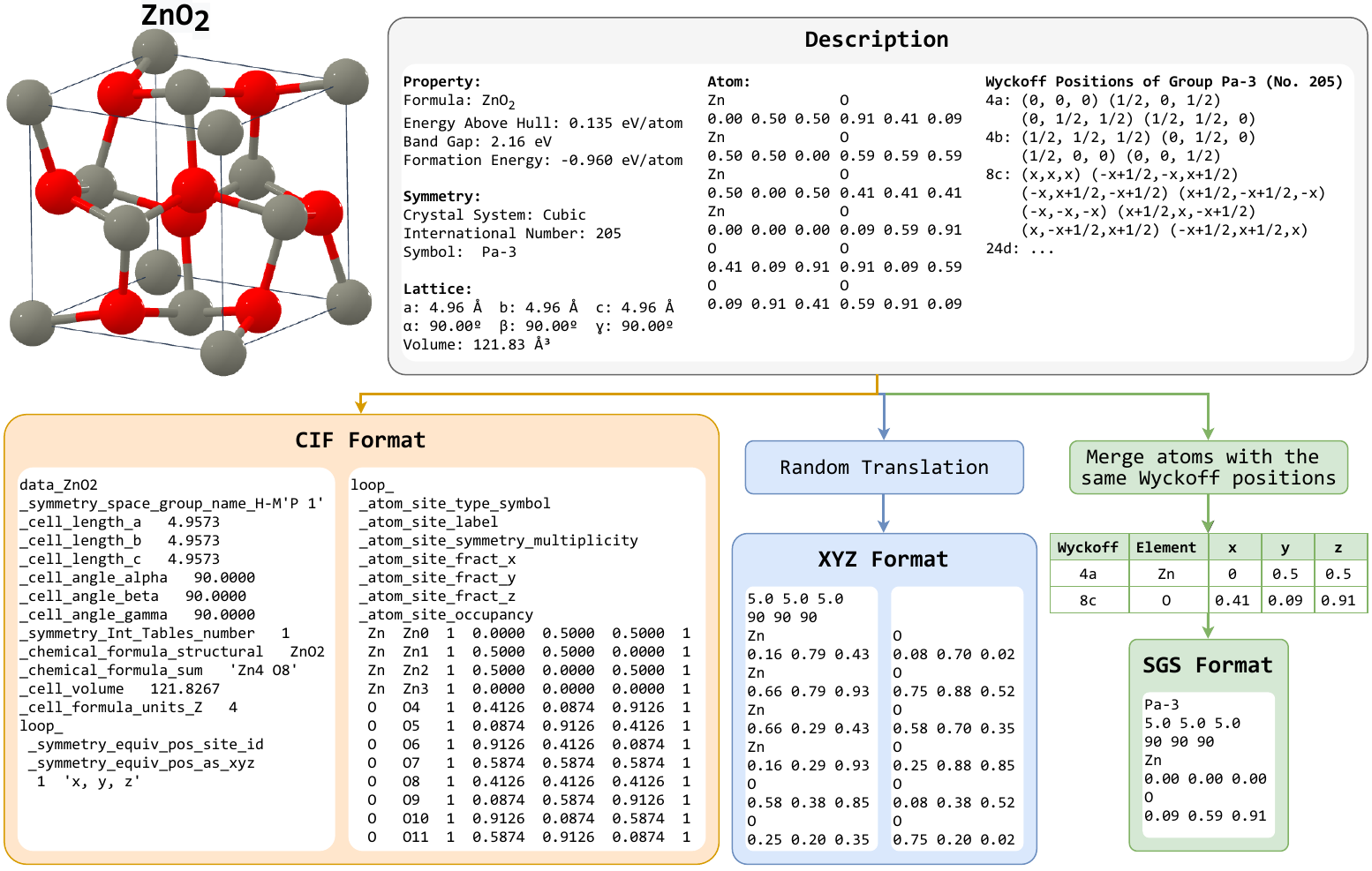}
    \vspace{-1.5em}
    \caption{Comparison of three methods (CIF, XYZ, SGS(ours)) for converting crystal structures to text.} \label{SGSImg}
    \vspace{-1.0em}
\end{figure}

To address the challenge of converting complex unit cell structures into text formats suitable for language models, we propose a novel space-group based crystal tokenization method (SGS) that reduces the complexity of crystal symmetry modeling for LLMs. Existing formats, such as CIF~\cite{hall1991crystallographic} and XYZ~\cite{LMCH}, exhibit notable limitations: CIF files, as highly formatted documents, include complex structures and a large number of specialized tokens thereby increasing the complexity of fine-tuning LLMs. Meanwhile, the XYZ format represents the fractional coordinates of all atoms within the unit cell, 
requiring the model to implicitly learn intricate symmetry relationships among atomic positions without structural guidance.
Therefore, these limitations exacerbate the difficulty of crystal generation tasks and hinder model performance.

Our method simplifies the crystal structure text by leveraging the concept of Wyckoff positions~\cite{LIPSON1949} in crystallography. A Wyckoff position is defined as a set of points whose site symmetry groups are all conjugate subgroups of one another. The space group uniquely determines the types of Wyckoff positions present in a crystal. Therefore, given the space group, atoms of the same element occupying the same Wyckoff position can be represented by a single atom. This decomposes the task of predicting the fractional 3D coordinates of all atoms in the unit cell into two components: modeling the correspondence between the space group and Wyckoff positions, and predicting the Wyckoff positions of atoms within the unit cell.

Specifically, our space-group based crystal structure text consists of three components: the space group symbol, the lattice parameters, and the element symbols and fractional coordinates for atoms at each Wyckoff position. An example of crystal string formatting is shown in Fig. \ref{SGSImg}. By replacing multiple atoms sharing the same Wyckoff position with a single representative atom, 
our method reduces the number of atomic coordinates that need to be generated and eliminates the need to enforce strict symmetry constraints during the atomic coordinate generation process, thereby lowering the difficulty of modeling crystallographic symmetry. This transformation simplifies the modeling of crystal symmetry, enabling LLMs to focus on key structural features, thereby reducing the complexity of the generation task and improving generation performance.

\subsection{Condition-Structure Aware Hybrid Crystal Instruction Tuning} \label{HIns}
Following the crystal tokenization process, the subsequent step is to construct an instruction tuning set $S_D$ for the crystal generation tasks.
The tuning set consists of two components: the zero-shot instruction tuning set $S_{D_z}$ and the few-shot instruction tuning set $S_{D_f}$.
In accordance with the standard protocol for instruction fine-tuning \cite{CrystalLLM, zhang2023moleculegpt}, the zero-shot crystal generation instruction set $S_{D_z}$ can be constructed from the given dataset $D$ using the following prompt template $T_z=\{Q, R\}$, where $Q$ represents the query and $R$ denotes the response, providing the necessary contextual information for the task.:
\begin{Verbatim}[breaklines=true,commandchars=\\\{\}]
### Instruction: Below is a description of a bulk material. \textcolor{blue}{[Condition Description]}. Generate the space group symbol, a description of the lengths and angles of the lattice vectors and then the element type and coordinates for each atom within the lattice:
### Response: \textcolor{red}{[Crystal String]}.
\end{Verbatim}

We have shown how instruction fine-tuning can guide LLMs in zero-shot crystal generation tasks. However, the lack of contextual learning in zero-shot settings prevents the model from fully leveraging its powerful ICL capabilities, thus limiting its potential in crystal generation. To address this limitation, we propose a few-shot instruction design method that incorporates crystal structures from the target generation domain into the prompt. The key idea is to use a small set of target-domain crystal structures as demonstrations to guide the model in generating similar crystal structures, thereby enhancing its conditional generation capabilities.

To achieve this, given a set of crystal properties $S=\{s_1, s_2, \cdots, s_n\}$ serving as generation conditions, we explore three different strategies for selecting $K$ representative crystals:

\textbf{Condition-based selection}. 
The first strategy filters the dataset based on the specified properties, ensuring that the selected crystals meet the given generation conditions $S$.
For chemical formula, anonymized representations (e.g., expressing $\text{CaTiO}_3$ as $\text{ABC}_3$) are used to generalize composition-based filtering. 
For discrete properties such as space group, crystals are selected by exact property matching. For continuous properties like band gap, we rank the crystals in ascending order of the absolute difference between their property values and the target condition, selecting those closest to the desired value.

\textbf{Structure-based selection}. In contrast, this strategy does not rely on explicit property constraints but instead retrieves structurally similar crystals from the dataset. A crystal is randomly chosen as an anchor crystal, and the CrystalNN fingerprint \cite{CrystalNN} is computed for all other crystals. The top $K-1$ crystals with the smallest euclidean distance to the anchor crystal are then selected, ensuring that structurally similar $K$ crystals serve as few-shot examples.

\textbf{Condition-Structure based selection}. 
To balance property consistency and structural similarity, this strategy combines the strengths of both approaches.
We first filter the dataset to obtain crystals that meet the specified conditions and randomly select a crystal as anchor. 
Then, we retrieve the $K-1$ crystals with the highest structural similarity to the anchor based on euclidean distance in the CrystalNN fingerprint space.
The final few-shot example set consists of the anchor crystal and its $K-1$ nearest neighbors, facilitating both property-conditioned and structure-aware generation.

By integrating the above three selection strategies, we construct the few-shot example set and extend the zero-shot instruction template to obtain the few-shot template $T_f$:
\begin{Verbatim}[breaklines=true,commandchars=\\\{\}]
### Instruction: Below is three description of bulk materials. 
### First Example:
### \textbf{[Condition Description-1]}
### \textbf{[Crys Str-1]}
### ...
### \textcolor{blue}{[Condition Description]}. Based on the three examples provided, generate the space group symbol, a description of the lengths and angles of the lattice vectors, along with the element type and coordinates for each atom within the lattice: 
### Response: \textcolor{red}{[Crystal String]}.
\end{Verbatim}

Given the constructed condition-structure aware instruction template $T_f$, the few-shot crystal generation instruction set $S_{D_f}$ can be generated by applying $T_f$ to each sample in $D$. The fine-tuning instruction set $S_D$ is obtained by combining the zero-shot instruction set $S_{D_z}$ with the few-shot instruction set $S_{D_f}$. The pre-trained LLM can then be fine-tuned by optimizing the following training loss:
\begin{equation}
\mathcal{L}(\theta) = \sum_{(Q_i, R_i)\in S_D} - \log f_\theta (R_i|Q_i),
\end{equation}
where $f_\theta$ represents the pre-trained LLM parameterized by $\theta$. In practice, $f_\theta$ is initialized using Llama2-7b-chat, and LoRA \cite{LoRA} is employed to accelerate the training process. Further details can be found in Appendix \ref{sec:appendix_param}.

\subsection{Multi-Task Crystal Instruction Tuning: Property Prediction Auxiliary Task}\label{Inf}

In the previous section, we introduce a condition-structure aware hybrid crystal instruction tuning framework that enhances ICL ability by integrating hybrid example selection methods. However, the model still lacks explicit supervision for learning the intrinsic mapping between crystal structures and their physicochemical properties, which is critical for generating accurate and meaningful crystals.
To address this limitation, we propose a multi-task crystal instruction tuning strategy that incorporates the property prediction auxiliary task. In addition to the primary crystal generation task, this auxiliary task trains the model to predict key crystal properties based on the crystal structure text. We define the property prediction template $T_p$ as follows:
\begin{Verbatim}[breaklines=true,commandchars=\\\{\}]
### Instruction: Below is a partial description of a bulk material where the \textbf{[Property]} has been replaced with the string "[MASK]":
### The \textbf{[Property]} is [MASK].
### \textcolor{blue}{[Crys Str]}
### Generate the \textbf{[Property]} that could replace [MASK] in the bulk material:
### Response: \textcolor{red}{[Property Value]}.
\end{Verbatim}

By jointly optimizing both crystal generation and property prediction tasks, the model learns to internalize structural patterns and their corresponding physical attributes, improving both generation accuracy and property consistency. 

\section{Experiment}

\begin{table*}[ht]
    \centering
    \resizebox{\textwidth}{!}{
    \renewcommand{\arraystretch}{0.9}
    \begin{tabular}{c|c|cc|cc|cc|cc}
        \toprule
        \multirow{3}{*}{Dataset} & \multirow{3}{*}{Method} & \multicolumn{8}{c}{\textbf{Success Rate}} \\
        ~ & ~ & \multicolumn{2}{c|}{Pretty Formula} & \multicolumn{2}{c|}{Space Group} & \multicolumn{2}{c|}{Formation Energy} & \multicolumn{2}{c}{Band Gap} \\ 
        ~ & ~ & Mean & Std. & Mean & Std. & Mean & Std. & Mean & Std. \\ \midrule
\multirow{6}{*}{MP20} & CrystalLLM (XYZ) & 0.9394 & 0.0099 & 0.0640 & 0.0078 & 0.8475 & 0.0169 & 0.6637 & 0.0129 \\
~ & CrystalICL (XYZ) 0-Shot & 0.9578 & 0.0067 & 0.0868 & 0.0098 & 0.8751 & 0.0048 & 0.6655 & 0.0233 \\
~ & CrystalICL (XYZ) 3-Shot & \textbf{0.9906} & 0.0050 & \textbf{0.0886} & 0.0151 & \textbf{0.9125} & 0.0072 & \textbf{0.7087} & 0.0165 \\ \cmidrule{2-10}
~ & CrystalLLM (SGS) & 0.4513 & 0.0218 & 0.8726 & 0.0097 & 0.7984 & 0.0144 & 0.6373 & 0.0159 \\
~ & CrystalICL (SGS) 0-Shot & 0.7218 & 0.0135 & 0.9881 & 0.0050 & 0.9049 & 0.0170 & 0.7023 & 0.0146 \\
~ & CrystalICL (SGS) 3-Shot & \textbf{0.8868} & 0.0077 & \textbf{0.9908} & 0.0033 & \textbf{0.9392} & 0.0094 & \textbf{0.7453} & 0.0263 \\ \midrule
\multirow{6}{*}{MP30} & CrystalLLM (XYZ) & 0.9699 & 0.0019 & 0.0799 & 0.0087 & 0.8297 & 0.0091 & 0.6732 & 0.0211 \\
~ & CrystalICL (XYZ) 0-Shot & 0.9536 & 0.0066 & 0.0926 & 0.0158 & 0.8485 & 0.0093 & 0.6767 & 0.0273 \\
~ & CrystalICL (XYZ) 3-Shot & \textbf{0.9922} & 0.0028 & \textbf{0.1083} & 0.0089 & \textbf{0.9461} & 0.0056 & \textbf{0.7454} & 0.0139 \\ \cmidrule{2-10}
~ & CrystalLLM (SGS) & 0.5008 & 0.0253 & 0.9006 & 0.0098 & 0.8030 & 0.0177 & 0.6687 & 0.0220 \\
~ & CrystalICL (SGS) 0-Shot & 0.7162 & 0.0118 & 0.9827 & 0.0052 & 0.8642 & 0.0143 & 0.6782 & 0.0053 \\
~ & CrystalICL (SGS) 3-Shot & \textbf{0.9641} & 0.0075 & \textbf{0.9956} & 0.0028 & \textbf{0.9789} & 0.0043 & \textbf{0.7943} & 0.0098 \\
        \bottomrule
    \end{tabular}
    }
    \vspace{-0.5em}
    \caption{The conditional sample performance on MP20 and MP30.}\label{CSInD_C4}
    \vspace{-2.0em}
\end{table*}

In our experiments, we aim to address five key research questions: 
\textbf{RQ1}: Can CrystalICL effectively inherit the ICL capabilities of LLMs and leverage limited examples in the prompt to
improve performance in the conditional crystal generation task?
\textbf{RQ2}: Can CrystalICL effectively utilize its ICL capabilities to achieve cross-domain conditional generation?
\textbf{RQ3}: How does CrystalICL perform in the unconditional crystal generation task compared to existing baselines?
\textbf{RQ4}: How do different types of instructions in the tuning set affect CrystalICL's ICL ability?
\textbf{RQ5}: How do different example selection strategies during inference affect the performance of CrystalICL?

\subsection{Experimental Setup}

\textbf{Datasets.} 
We evaluate the conditional generation task on four crystal generation datasets: MP20 \cite{MP20}, MP30 \cite{CrystalLLM}, P5 \cite{P5}, and C24 \cite{C24}. For the unconditional generation task, following the previous work \cite{CDVAE, DiffCSP}, we evaluate on MP20, P5, and C24.
Detailed dataset information is provided in Appendix \ref{sec:appendix_dataset}. 
For the conditional generation task, we employ the Success Rate as the evaluation metric. For the unconditional generation task, we evaluate performance across three key aspects: Validity, Coverage and Property Distribution. Further details can be found in Appendix \ref{sec:appendix_metric}.

\textbf{Baselines.} For the conditional crystal generation task, we use CrystalLLM \cite{CrystalLLM} as the baseline model. For the unconditional crystal generation task, we select CDVAE \cite{CDVAE}, DiffCSP \cite{DiffCSP}, and CrystalLLM as the baseline models. 

\subsection{Conditional Generation Evaluation}

\begin{figure}[t]
    \centering
    \includegraphics[width=\columnwidth]{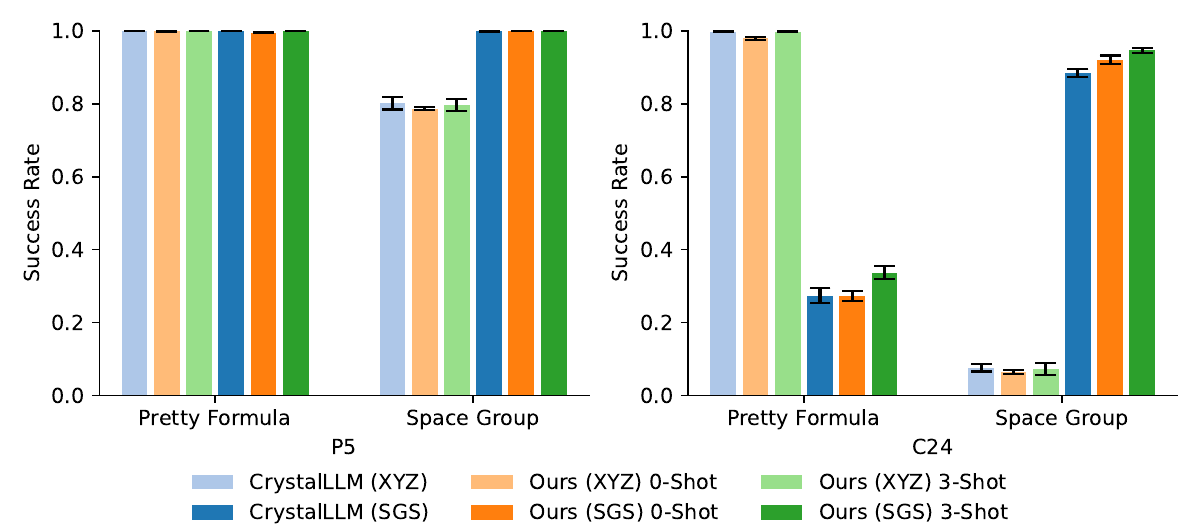}
    \vspace{-1.5em}
    \caption{The conditional sample performance on P5 and C24.} \label{CSInD}
    \vspace{-2.0em}
\end{figure}

To address RQ1, we conduct evaluations of conditional generation task across four datasets. 
Specifically, we select the properties of crystals from the test set as generation conditions, and use examples from the training set as demonstrations in the few-shot prompt. Further details can be found in Appendix \ref{sec:appendix_metric}.
Each fine-tuned model is evaluated through five iterations of 1,000-sample testing, and Tab. \ref{CSInD_C4} and Fig. \ref{CSInD} summarize the mean performance metrics along with their standard deviation across these iterations. Based on the experimental results, we summarize two key conclusions:

\textbf{CrystalICL is effective and reliable.}
As shown in Tab. \ref{CSInD_C4} and Fig. \ref{CSInD}, CrystalICL exhibits superior performance in conditional crystal generation task across various crystal structure text formats and datasets of different domains and scales. 
In contrast to CrystalLLM, which experiences a significant performance drop in zero-shot scenarios for chemical formula generation using SGS-format crystal structure text, our CrystalICL shows a smaller decline, demonstrating its robustness across different formats of crystal structure text. 
Furthermore, CrystalICL shows a marked improvement in performance in few-shot scenarios compared to zero-shot scenarios, effectively validating its successful inheritance of the ICL capabilities of LLMs.

\textbf{SGS proves to be an effective method for crystal tokenization.} 
The results indicate that, compared to XYZ-format crystal structure text, SGS significantly enhances LLMs' ability to generate crystals conditioned on the space group, while performance for chemical-formula based tasks is slightly reduced.
This highlights the effectiveness of our space-group based crystal tokenization method in simplifying the complexity of modeling crystal symmetry. Furthermore, on both the MP20 and MP30 datasets, the SGS-based approach outperforms in crystal physicochemical property conditioned generation tasks, demonstrating the importance of crystal symmetry in modeling the relationship between crystal structures and their properties. Additionally, the use of XYZ-format crystal structure text results in notably poor performance in space-group conditioned generation across multiple datasets, supporting our view raised in Sec. \ref{SGS} that including the 3D coordinates of all atoms in the crystal structure text exacerbates the challenge of modeling crystal symmetry in LLMs.

\subsection{Cross-domain Conditional Generation Evaluation}

To address RQ2, we design a cross-domain conditional generation scenario for evaluation. The model is trained on the MP20 dataset, while testing is conducted on randomly selected crystals from the test sets of P5 and C24, using their properties as generation conditions. 

\begin{figure}[t]
    \centering
    \includegraphics[width=\columnwidth]{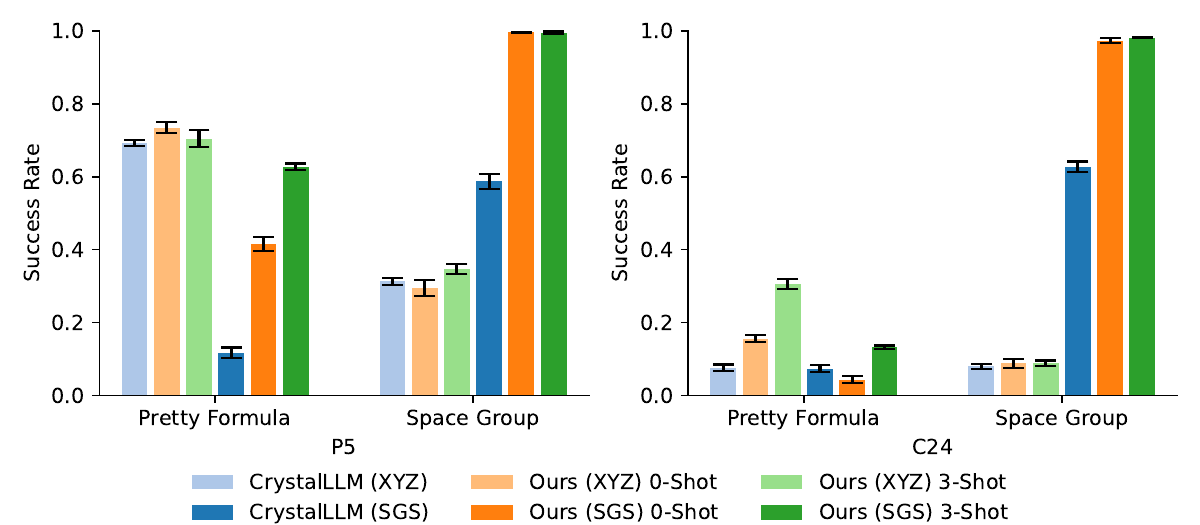}
    \vspace{-1.5em}
    \caption{The cross-domain conditional generation performance on P5 and C24.} \label{CSOOD}
    \vspace{-1.5em}
\end{figure}

\textbf{With the use of SGS, CrystalICL effectively leverages its ICL capabilities to achieve cross-domain conditional generation.} As shown in Fig. \ref{CSOOD}, the performance of the cross-domain conditional generation task experiences a significant decline when compared to the domain-specific conditional generation task shown in Fig. \ref{CSInD}. This is primarily due to the considerable differences between the MP20, P5, and C24 datasets. Specifically, all crystals in the P5 dataset share the same chemical formula $\text{ABX}_3$, and only four space groups are involved. In the C24 dataset, all crystals consist solely of carbon atoms. Moreover, most of the crystals in P5 and C24 do not exist in reality, making it challenging to transfer knowledge from the MP20 dataset to these datasets. 
However, it is encouraging to observe that when using SGS-format crystal structure text, CrystalICL demonstrates remarkable cross-domain performance in space-group conditioned generation, with almost no performance degradation compared to domain-specific conditional generation task. Additionally, when compared to CrystalLLM, CrystalICL exhibits a smaller performance decline in the chemical-formula conditioned generation task, showcasing its stronger generalization ability.

\subsection{Unconditional Generation Evaluation}

To answer RQ3, we use the unconditional generation prompt to sample 10,000 structures from each fine-tuned model and attempt to parse them into CIF files based on the generated samples. If a sampled string cannot be parsed as a valid CIF, the sample is rejected and re-sampled.

\textbf{CrystalICL demonstrates a superior ability to capture the relationship between crystal structures and their properties.} 
As shown in Tab. \ref{UC}, we evaluate the performance of unconditional crystal generation task across multiple datasets. 
The results indicate that, compared to CrystalLLM, CrystalICL generates crystal structures whose property distributions exhibit a much closer alignment with those in the training sets across all three datasets. 
This demonstrates the effectiveness of CrystalICL in learning both crystal structure and property distributions. It also validates the impact of incorporating few-shot crystal generation instructions into the instruction tuning set for modeling crystal distributions.
Additionally, SGS-format crystal structure text reduce the distributional differences in density and formation energy across all three datasets, demonstrating the effectiveness of SGS in modeling the relationship between crystal structures and properties. Particularly, for the C24 dataset, which consists entirely of carbon-based materials, SGS significantly alleviates the difficulty for LLMs in modeling crystal structure distributions, leading to a substantial improvement in precision and recall on this dataset.

\begin{table}[t]
    \centering
    \resizebox{\columnwidth}{!}{
    \begin{tabular}{c|c|ccc|cc|ccc}
        \toprule
        \multirow{2}{*}{Dataset} & \multirow{2}{*}{Method} & \multicolumn{3}{c|}{\textbf{Validity Check $\uparrow$}} & \multicolumn{2}{c|}{\textbf{Coverage  $\uparrow$}} & \multicolumn{3}{c}{\textbf{Property Distribution  $\downarrow$}}
        \\
        ~ & ~ & Composition & Structural & Valid & Recall & Precision & wdist($\rho$) & wdist($E$) & wdist($N_{el}$) \\ \midrule
\multirow{6}{*}{MP20} & CDVAE & 0.8514 & 0.9999 & 0.8514 & 0.9896 & 0.9946 & 0.6445 & 0.2617 & 1.1567 \\
~ & DiffCSP & 0.8182 & 0.9983 & 0.8172 & 0.9957 & 0.9967 & 0.1907 & 0.1394 & 0.5703 \\
~ & CrystalLLM(XYZ) & 0.9019 & 0.9630 & 0.8697 & 0.9839 & 0.9931 & 1.3315 & 0.4503 & 0.1811 \\
~ & CrystalICL(XYZ) & 0.8922 & 0.9792 & 0.8747 & 0.9840 & 0.9949 & 1.2175 & 0.3818 & 0.1756 \\
~ & CrystalLLM(SGS) & 0.8433  & 0.9570  & 0.8144  & 0.9944  & 0.9847  & 0.8356  & 0.3544  & 0.1743 \\
~ & CrystalICL(SGS) & 0.8655 & 0.9859 & 0.8555 & 0.9949 & 0.9926 & 0.6039 & 0.2568 & 0.1359 \\ \midrule
\multirow{6}{*}{P5} & CDVAE & 0.9841 & 1.0000 & 0.9841 & 0.9897 & 0.9852 & 0.0664 & 0.0474 & 0.1350 \\
~ & DiffCSP & 0.9848 & 0.9999 & 0.9845 & 0.9947 & 0.9820 & 0.0462 & 0.0532 & 0.0301 \\
~ & CrystalLLM(XYZ) & 0.9896 & 1.0000 & 0.9896 & 0.9915 & 0.9856 & 0.1755 & 0.0448 & 0.0196 \\
~ & CrystalICL(XYZ) & 0.9940 & 1.0000 & 0.9940 & 0.9905 & 0.9898 & 0.2269 & 0.0284 & 0.0200 \\
~ & CrystalLLM(SGS) & 0.9895 & 1.0000 & 0.9895 & 0.9918 & 0.9880 & 0.1443 & 0.0155 & 0.0333 \\
~ & CrystalICL(SGS) & 0.9916 & 1.0000 & 0.9916 & 0.9908 & 0.9872 & 0.1692 & 0.0089 & 0.0644 \\ \midrule
\multirow{6}{*}{C24} & CDVAE & 1.0000 & 1.0000 & 1.0000 & 0.9990 & 0.8416 & 0.1497 & 0.2206 & - \\
~ & DiffCSP & 1.0000 & 1.0000 & 1.0000 & 0.9990 & 0.9642 & 0.0548 & 0.0415 & - \\
~ & CrystalLLM(XYZ) & 1.0000 & 0.8893 & 0.8893 & 0.2182 & 0.0005 & 0.0691 & 30.8053 & - \\
~ & CrystalICL(XYZ) & 1.0000 & 0.9282 & 0.9282 & 0.5458 & 0.0052 & 0.0395 & 29.4212 & - \\
~ & CrystalLLM(SGS) & 1.0000 & 0.9574 & 0.9574 & 0.9916 & 0.7965 & 0.0707 & 3.0564 & - \\
~ & CrystalICL(SGS) & 1.0000 & 0.9669 & 0.9669 & 0.9921 & 0.8928 & 0.0593 & 1.3061 & - \\
        \bottomrule
    \end{tabular}
    }
    \vspace{-0.5em}
    \caption{The unconditional sample performance.}\label{UC}
    \vspace{-1.5em}
\end{table} 

\subsection{Effect of Instruction Types in the Tuning Set} \label{Ablation}

To investigate RQ4, we conduct an ablation study to analyze how different instruction types affect conditional crystal generation performance. Specifically, to better align with real-world ICL scenarios, we train on the MP20 and select demonstrations from the MP30 (an extension of MP20), examining the influence of various instruction types on performance during the training process, with the results presented in Tab. \ref{TrainMethodMP20SGS}.
In this study, C, F, and CF correspond to the three demonstration selection strategies introduced in Sec. \ref{HIns}, Rand represents the random selection strategy, and noAux refers to the removal of the property prediction auxiliary instruction set described in Sec. \ref{Inf}. 

\textbf{Hybrid instruction tuning effectively enhances the capabilities of CrystalICL across various scenarios.} Experimental results indicate that variants using a single example selection strategy during fine-tuning, such as F, CF, and C, show poorer performance in zero-shot scenarios. However, in the few-shot setting, the performance remains consistent with that in the zero-shot scenario, demonstrating that randomly example selection strategy leads to failing to derive task-relevant information from the demonstrations, thus losing ICL capability. Additionally, removing the property prediction instructions from the tuning set results in a decline in zero-shot performance, highlighting the importance of explicitly designing instructions to guide the model in learning the relationship between crystal structures and their properties.

\begin{table}[t]
    \centering
    \resizebox{\columnwidth}{!}{
    \begin{tabular}{c|c|cc|cc|cc|cc}
        \toprule
        \multirow{3}{*}{Scenario} & \multirow{3}{*}{Method} & \multicolumn{8}{c}{\textbf{Success Rate}} \\
        ~ & ~ & \multicolumn{2}{c|}{Pretty Formula} & \multicolumn{2}{c|}{Space Group} & \multicolumn{2}{c|}{Formation Energy} & \multicolumn{2}{c}{Band Gap} \\
        ~ & ~ & Mean & Std. & Mean & Std. & Mean & Std. & Mean & Std. \\ \midrule
\multirow{6}{*}{0-Shot} & Rand & 0.7364 & 0.0133 & 0.9916 & 0.0026 & 0.9082 & 0.0147 & 0.7006 & 0.0247 \\
~ & F & 0.2966 & 0.0039 & 0.6446 & 0.0169 & 0.8039 & 0.0170 & 0.6239 & 0.0149 \\
~ & CF & 0.6854 & 0.0081 & 0.9850 & 0.0039 & 0.8836 & 0.0186 & 0.7077 & 0.0126 \\
~ & C & 0.6620 & 0.0180 & 0.9788 & 0.0035 & 0.8917 & 0.0082 & 0.6941 & 0.0213 \\
~ & noAux & 0.4167 & 0.0227 & 0.3740 & 0.0178 & 0.8370 & 0.0047 & 0.6742 & 0.0156 \\ \cmidrule{2-10}
~ & CrystalICL & 0.7390 & 0.0313 & 0.9904 & 0.0040 & 0.9104 & 0.0084 & 0.7017 & 0.0218 \\ \midrule
\multirow{6}{*}{3-Shot} & Rand & 0.7430 & 0.0140 & 0.9883 & 0.0023 & 0.9083 & 0.0103 & 0.6982 & 0.0180 \\
~ & F & 0.5425 & 0.0146 & 0.8436 & 0.0154 & 0.8144 & 0.0105 & 0.6473 & 0.0145 \\
~ & CF & 0.4878 & 0.0062 & 0.3681 & 0.0202 & 0.8528 & 0.0115 & 0.6991 & 0.0156 \\
~ & C & 0.9340 & 0.0075 & 0.9954 & 0.0016 & 0.9669 & 0.0073 & 0.7789 & 0.0214 \\
~ & noAux & 0.9347 & 0.0078 & 0.9959 & 0.0010 & 0.9647 & 0.0090 & 0.7765 & 0.0253 \\ \cmidrule{2-10}
~ & CrystalICL & 0.9214 & 0.0104 & 0.9948 & 0.0033 & 0.9685 & 0.0087 & 0.7687 & 0.0146 \\
        \bottomrule
    \end{tabular}
    }
    \vspace{-0.5em}
    \caption{Impact of different instruction types  in the 3-Shot Setting.}\label{TrainMethodMP20SGS}
    \vspace{-1.0em}
\end{table}

\subsection{Effect of Example Selection Strategies During Inference}

\begin{table}[t]
    \centering
    \resizebox{\columnwidth}{!}{
    \begin{tabular}{c|cc|cc|cc|cc}
        \toprule
        \multirow{3}{*}{Method} & \multicolumn{8}{c}{\textbf{Success Rate}} \\
        ~ & \multicolumn{2}{c|}{Pretty Formula} & \multicolumn{2}{c|}{Space Group} & \multicolumn{2}{c|}{Formation Energy} & \multicolumn{2}{c}{Band Gap} \\
        ~ & Mean & Std. & Mean & Std. & Mean & Std. & Mean & Std. \\ \midrule
FR & 0.7284 & 0.0131 & 0.9807 & 0.0041 & 0.9064 & 0.0079 & 0.6957 & 0.0173 \\
F & 0.7273 & 0.0116 & 0.9846 & 0.0014 & 0.9020 & 0.0061 & 0.6981 & 0.0246 \\
R & 0.7441 & 0.0147 & 0.9871 & 0.0046 & 0.9107 & 0.0094 & 0.7120 & 0.0167 \\
CFR & 0.8417 & 0.0057 & 0.9925 & 0.0041 & 0.9457 & 0.0047 & 0.7237 & 0.0124 \\
CF & 0.8445 & 0.0106 & 0.9931 & 0.0026 & 0.9346 & 0.0061 & 0.7327 & 0.0036 \\
CR & 0.9024 & 0.0046 & 0.9950 & 0.0028 & 0.9532 & 0.0084 & 0.7655 & 0.0111 \\
C & 0.9214 & 0.0104 & 0.9948 & 0.0033 & 0.9685 & 0.0087 & 0.7687 & 0.0146 \\
        \bottomrule
    \end{tabular}
    }
    \vspace{-0.75em}
    \caption{Impact of different example selection strategies in the 3-Shot Setting.}\label{3ShotSampleMethodMP20SGS}
    \vspace{-1.5em}
\end{table}

To answer RQ5, we conduct experiments to evaluate the impact of different example selection strategies and the number of examples. The experimental setup follows the same configuration as described in Sec. \ref{Ablation}. 
The results of these strategies under the 3-shot setting are presented in Tab. \ref{3ShotSampleMethodMP20SGS}.
Specifically, C, F, and CF correspond to the three example selection strategies introduced in Sec. \ref{HIns}, while R refers to the random selection of examples or the shuffling of the selected examples order.
Based on the experimental results in Tab. \ref{3ShotSampleMethodMP20SGS} and Tab. \ref{NShotMP20SGS}, we summarize two conclusive findings as follows:

\textbf{Providing appropriate demonstrations is crucial for crystal generation.} The results in Tab. \ref{3ShotSampleMethodMP20SGS} demonstrate that shuffling the order of examples reduces the success rate of conditional generation, confirming the effectiveness of using generation conditions and CrystalNN fingerprints as criteria for example selection. Moreover, the condition-based prompt construction strategy achieves the best performance, highlighting the importance of providing similar examples in conditional crystal generation task.

\textbf{The number of demonstrations has an insignificant impact on crystal generation task.} Tab. \ref{NShotMP20SGS} presents the influence of the number of examples on conditional generation performance, evaluating both random selection and condition-based selection strategies for prompt construction. The results indicate that the number of examples selected during generation has no significant impact on the model's cross-domain conditional generation capability.

\begin{table}[t]
    \centering
    \resizebox{\columnwidth}{!}{
    \begin{tabular}{c|cc|cc|cc|cc}
        \toprule
        \multirow{3}{*}{Method} & \multicolumn{8}{c}{\textbf{Success Rate}} \\
        ~ & \multicolumn{2}{c|}{Pretty Formula} & \multicolumn{2}{c|}{Space Group} & \multicolumn{2}{c|}{Formation Energy} & \multicolumn{2}{c}{Band Gap} \\
        ~ & Mean & Std. & Mean & Std. & Mean & Std. & Mean & Std. \\ \midrule
1Shot-R & 0.7360 & 0.0177 & 0.9866 & 0.0041 & 0.9026 & 0.0066 & 0.7070 & 0.0160 \\
2Shot-R & 0.7310 & 0.0084 & 0.9865 & 0.0066 & 0.9072 & 0.0054 & 0.7023 & 0.0120 \\
3Shot-R & 0.7441 & 0.0147 & 0.9871 & 0.0046 & 0.9107 & 0.0094 & 0.7120 & 0.0167 \\ \midrule
1Shot-C & 0.9376 & 0.0039 & 0.9946 & 0.0024 & 0.9731 & 0.0045 & 0.8009 & 0.0235 \\
2Shot-C & 0.9410 & 0.0119 & 0.9954 & 0.0020 & 0.9683 & 0.0036 & 0.7777 & 0.0068 \\
3Shot-C & 0.9214 & 0.0104 & 0.9948 & 0.0033 & 0.9685 & 0.0087 & 0.7687 & 0.0146 \\
        \bottomrule
    \end{tabular}
    }
    \vspace{-0.5em}
    \caption{Impact of different shot numbers.}\label{NShotMP20SGS}
    \vspace{-1.5em}
\end{table}

\section{Related work}

\subsection{Equivariant Diffusion model-based Crystal Generation Methods}
Periodicity and symmetry are fundamental characteristics of crystals, which have a decisive impact on their physical properties. Therefore, ensuring the SE(3) equivariance
of the crystal generation process is crucial for crystal generation task \cite{SurveyGGNN}. To address this challenge, equivariant diffusion models \cite{CDVAE, DiffCSP, DiffCSPPP} have emerged as a leading method for generating stable crystal materials in recent years, owing to their ability to harness the physical symmetries inherent in periodic material structures. 

\subsection{Language model-based Crystal Generation Methods}
In recent years, LLMs trained on large-scale unsupervised corpora have demonstrated unprecedented powerful capabilities across various tasks, which has stimulated researchers' interest in the potential of language models in learning effective "world models" for crystal chemistry. Several studies \cite{CrystalLM, CrystalLLM} have begun to explore the capabilities of language models in the field of crystal generation. 

\section{Conclusion}

In this work, we propose CrystalICL, a crystal generation model designed to effectively inherit the in-context learning (ICL) generalization ability of LLMs.
CrystalICL leverages the Wyckoff position to simplify the modeling of crystal symmetry and constructs an instruction fine-tuning set that harnesses the ICL capabilities of LLMs for crystal generation tasks.
Extensive experiments on four crystal generation benchmarks demonstrate the ICL capabilities of CrystalICL on conditional and unconditional generation tasks across domains.

\bibliography{custom}

\appendix

\section{Failure Example}
\label{sec:appendix_mov}
The following is an example of a failed 0-shot generation by GPT-3.5 Turbo:

\begin{Verbatim}[breaklines=true,commandchars=\\\{\}]
### \textcolor{blue}{Instruction:}
### Example:
### a b c
### alpha beta gamma 
### atom
### x y z
### atom
### x y z
### Below is a description of a bulk material. The chemical formula is SnReN3. The spacegroup number is 221. Generate a description of the lengths and angles of the lattice vectors and then the element type and coordinates for each atom within the lattice. The answer should contain only the element symbol and the number, and should be formatted as given in the example:
### \textcolor{red}{Response:}
### Description of lattice vectors:
### a = 5.67 Å
### b = 5.67 Å
### c = 5.67 Å
### alpha = 90°
### beta = 90°
### gamma = 90°
### 
### Element type and coordinates:
### Sn 0.00 0.00 0.00
### Re 0.50 0.50 0.50
### N 0.25 0.25 0.25
### N 0.75 0.75 0.75
### N 0.00 0.50 0.00
### N 0.50 0.00 0.50
\end{Verbatim}

From the above failed example, it is evident that despite providing example structures in the prompt and specifying that the response should contain only element symbols and numbers, the output generated by GPT-3.5 Turbo still included special symbols such as '=', 'Å', and '°', as well as the English phrase "Description of lattice vectors:". This illustrates that a general pretrained large language model, without fine-tuning, struggles to adhere to the instructions and generate the desired crystal structure text.

\section{Details of Datasets}
\label{sec:appendix_dataset}
We selected four datasets with diverse domains and scales to comprehensively evaluate the model's performance. We apply the standard split as used in previous work \cite{CDVAE, CrystalLLM} for all of our experiments.

The MP20 dataset is derived from the Materials Project \cite{MP20} and contains 45,231 materials with varying structural and compositional characteristics, covering 89 elements and unit cells ranging from 1 to 20 atoms. Following the previous work \cite{CDVAE}, we only select structures with formation energy smaller than 2 eV/atom and energy above the hull smaller than 0.08 eV/atom. All materials in MP20 have been relaxed using density functional theory (DFT). The dataset analysis indicates that most materials exhibit thermodynamic stability and have been successfully synthesized in experiments.

The MP30 dataset \cite{CrystalLLM} is also derived from the Materials Project and comprises 127,609 crystal structures. The dataset spans a wide range of materials, with unit cells containing 1 to 30 atoms. Compared to MP20, MP30 includes a more extensive collection of structures, capturing a broader diversity of compositions. All crystal structures in MP30 have been relaxed using DFT to ensure reliable structural and energetic information, facilitating its application in crystal generation and property prediction tasks.

The Perov-5 (P5) dataset \cite{P5} consists of 18,928 perovskite materials that share the same structure but differ in chemical composition. It includes 56 elements, with each unit cell containing 5 atoms. Perovskite materials typically follow the general chemical formula $\text{ABX}_3$, where, in an ideal cubic structure, the $\text{A}$ atoms occupy the corner positions, the $\text{B}$ atoms are located at the body center, and the $\text{X}$ atoms are positioned at the face centers. Due to their wide range of applications, perovskites have attracted significant attention in photovoltaics, catalysis, and electronics. All structures in P5 have been relaxed using DFT. The resulting relaxed structures can deviate significantly from the ideal perovskite structures. Additionally, a significant portion of these materials are not thermodynamically stable, meaning they are prone to decomposition into more stable phases and are thus challenging to synthesize experimentally.

The Carbon-24 (C24) dataset comprises carbon structures generated through the ab initio random structure search (AIRSS) \cite{C24} method under a pressure of 10 GPa. While all materials in this dataset are composed exclusively of carbon, they exhibit considerable structural diversity. The dataset includes materials with unit cells containing between 6 and 24 atoms. 
Consistent with previous work \cite{CDVAE}, we retain only the 10\% of structures with the lowest energy per atom from the original dataset to create C24. All 10,153 structures in C24 have been optimized using DFT. The most stable structure under 10 GPa is diamond, whereas most remaining structures are thermodynamically unstable, although some may be kinetically stable. However, the majority of these structures are unlikely to be experimentally synthesizable.

\section{Evaluation Metrics}
\label{sec:appendix_metric}
For the evaluation of conditional crystal generation, we follow previous studies \cite{CSIJCAI, TGDMat} and adopt the success rate as the primary evaluation metric. 
On the MP30 dataset, we focus on the conditional generation on chemical formula, space group, formation energy, and band gap, as these properties can be easily validated or approximated. On the P5 and C24 datasets, we restrict the conditional constraints to chemical formula and space group due to limitations in the available data.
Chemical formula validation is performed by directly counting the atomic composition in the generated structures. Space group is determined using the SpacegroupAnalyzer module from the pymatgen \cite{pymatgen} library. Formation energy and band gap are estimated using MEGNet \cite{MEGNet} models trained on the Materials Project \cite{MP20} dataset. Specifically, for formation energy, a generated sample is considered successful if its formation energy shares the same sign as the input condition \cite{TGDMat}. For band gap, a sample is deemed valid if the absolute difference between the generated value and the input condition is less than 0.5 eV.
For the P5 and C24 datasets, since many materials in these datasets are derived from DFT calculations and do not exist in reality, their physicochemical properties may not be accurately estimated. Therefore, the evaluation for these datasets is limited to chemical formula and space group.

For the unconditional generation task, we follow prior work \cite{CDVAE, DiffCSP} and evaluate performance across three key aspects: validity metrics, coverage metrics, and property distribution metrics.
Validity metrics are categorized into structural validity and compositional validity. Structural validity is determined based on non-overlapping atomic radii, where overlap is defined as the distance between two atoms being less than half the sum of their radii. Compositional validity ensures that the generated structure has a net neutral charge, as only charge-neutral structures are considered valid.
Coverage metrics include recall and precision, which are computed based on CrystalNN fingerprints \cite{CrystalNN} and normalized Magpie fingerprints \cite{Magpie}. Recall measures how many ground-truth materials are correctly predicted, while precision assesses the quality of the generated materials.
Property distribution metrics are evaluated using the Wasserstein distance for three key properties: density $\rho$, formation energy per atom $E$, and the number of distinct element types $N_{el}$ within the unit cell. These metrics provide a comprehensive assessment of how well the generated crystal structures align with real-world material distributions.

\section{Baseline}
CrystalLLM \cite{CrystalLLM} finetunes Llama-2 on crystal structure texts in XYZ format, exploring both unconditional and conditional crystal generation tasks in zero-shot settings, thus highlighting the potential of LLMs in crystal generation. 
CDVAE \cite{CDVAE} proposes a VAE framework that first predicts the invariant lattice parameters and then generates the atom types and coordinates via a score-based decoder. DiffCSP \cite{DiffCSP} jointly generates the lattices and atom coordinates.

\section{Training Hyperparameters} \label{sec:appendix_param}
To efficiently fine-tune Llama2-7b-chat, this study employs the LoRA technique with a rank of 8, an alpha value of 32, and a dropout rate of 5e-2. The learning rate follows a cosine annealing schedule with an initial value of 5e-4.
During training, different batch sizes and epochs are used based on the dataset characteristics. For the MP20 and C24 datasets, a batch size of 1 is used with 10 training epochs. Due to the larger scale and extended training time required for the MP30 dataset, the batch size remains set to 1, while the number of training epochs is limited to 3. For the P5 dataset, where all crystal structures contain only five atoms per unit cell, a batch size of 4 is used with 10 training epochs to accelerate training.
In the inference stage, batch sizes are adjusted based on the specific generation task. The batch size is set to 6 for cross-domain conditional generation, 8 for conditional generation, and 32 for unconditional generation. Additionally, both the top-p sampling parameter and the temperature are set to 0.9.
All training and inference processes are conducted on a single Nvidia L40 48G GPU.

\section{Computational Cost of SGS Preprocessing}
To address concerns regarding the potential computational overhead introduced by the use of space-group based tokenization (SGS), we provide detailed runtime statistics for the conversion of CIF files to SGS format across various crystal datasets.

The SGS preprocessing involves the identification of space groups and Wyckoff positions for each crystal structure. This step is executed as a one-time offline transformation using pymatgen \cite{pymatgen} and spglib \cite{spglib}, accelerated by 32-process parallelization via the pandarallel library.

\begin{table}[h]
    \centering
    \resizebox{\columnwidth}{!}{
    \begin{tabular}{c|c|c}
        \toprule
        Dataset & Dataset Size & Conversion Time (s) \\
        \midrule
MP20 & 43k & 142.57 \\
MP30 & 144k & 442.73 \\
P5 & 19k & 20.94 \\
C24 & 10k & 12.93 \\
        \bottomrule
    \end{tabular}
    }
    \caption{Preprocessing time for CIF-to-SGS conversion using 32-process parallelism.}\label{preprocessing_time}
\end{table}

Table \ref{preprocessing_time} reports the wall-clock times for converting commonly used datasets on a server equipped with 4$\times$Intel(R) Xeon(R) Gold 5120 CPUs @ 2.20GHz.
As shown, SGS preprocessing is computationally efficient even for large-scale datasets. For example, the MP30 dataset, which includes over 140k crystals, can be converted in under 8 minutes. For smaller datasets, the conversion typically completes within seconds. Therefore, the computation of Wyckoff positions introduces minimal overhead and does not pose a bottleneck for practical usage.

\section{Details of Instructions}

Our instruction fine-tuning dataset consists of two components: the instruction section and the response section. The instruction section provides a description of the task, specifies the required output format, and may include optional few-shot examples. Depending on the specific fine-tuning task, the output can be either a textual representation of the crystal structure or a specific crystal property value. Below are examples illustrating different tasks from the four instruction fine-tuning datasets.

The following is an example of a 0-shot conditional generation instruction based on the MP20 dataset:

\begin{Verbatim}[breaklines=true,commandchars=\\\{\}]
### Instruction: Below is a description of a bulk material. \textcolor{blue}{The chemical formula is LiCuCO3. The energy above the convex hull is 0.0469. The spacegroup number is 67. The formation energy per atom is -1.681. The band gap is 1.7254.} Generate the space group symbol, a description of the lengths and angles of the lattice vectors and then the element type and coordinates for each atom within the lattice:
### Response: 
### \textcolor{red}{Cmme}
### \textcolor{red}{5.3 6.3 8.8}
### \textcolor{red}{90 90 90}
### \textcolor{red}{Li}
### \textcolor{red}{0.00 0.25 0.64}
### \textcolor{red}{Cu}
### \textcolor{red}{0.25 0.25 0.00}
### \textcolor{red}{C}
### \textcolor{red}{0.00 0.25 0.28}
### \textcolor{red}{O}
### \textcolor{red}{0.22 0.25 0.21}
### \textcolor{red}{O}
### \textcolor{red}{0.00 0.25 0.43}
\end{Verbatim}

The following is an example of a 3-shot conditional generation instruction from the P5 dataset:

\begin{Verbatim}[breaklines=true,commandchars=\\\{\}]
### Below is three description of bulk materials.
### First Example:
### \textbf{The chemical formula is BeBaO2F. The spacegroup number is 123.}
### \textbf{P4/mmm}
### \textbf{4.9 4.9 4.9}
### \textbf{90 90 90}
### \textbf{Ba}
### \textbf{0.50 0.50 0.50}
### \textbf{Be}
### \textbf{0.00 0.00 0.00}
### \textbf{O}
### \textbf{0.00 0.50 0.50}
### \textbf{F}
### \textbf{0.50 0.50 0.00}
### Second Example:
### \textbf{The chemical formula is ZrMoO2N. The spacegroup number is 123. }
### \textbf{P4/mmm}
### \textbf{4.0 4.0 4.0}
### \textbf{90 90 90}
### \textbf{Zr}
### \textbf{0.00 0.00 0.00}
### \textbf{Mo}
### \textbf{0.50 0.50 0.50}
### \textbf{N }
### \textbf{0.50 0.50 0.00}
### \textbf{O }
### \textbf{0.00 0.50 0.50}
### Third Example:
### \textbf{The chemical formula is NiNaO2F. The spacegroup number is 123. }
### \textbf{P4/mmm}
### \textbf{4.2 4.2 4.2}
### \textbf{90 90 90}
### \textbf{Na}
### \textbf{0.50 0.50 0.50}
### \textbf{Ni}
### \textbf{0.00 0.00 0.00}
### \textbf{O }
### \textbf{0.00 0.50 0.50}
### \textbf{F}
### \textbf{0.50 0.50 0.00}
### \textcolor{blue}{The chemical formula is MgAgO2F. The spacegroup number is 123.} Based on the three examples provided, generate the space group symbol, a description of the lengths and angles of the lattice vectors, along with the element type and coordinates for each atom within the lattice: 
### Response:
### \textcolor{red}{P4/mmm}
### \textcolor{red}{4.2 4.2 4.2}
### \textcolor{red}{90 90 90}
### \textcolor{red}{Mg}
### \textcolor{red}{0.00 0.00 0.00}
### \textcolor{red}{Ag}
### \textcolor{red}{0.50 0.50 0.50}
### \textcolor{red}{O}
### \textcolor{red}{0.00 0.50 0.50}
### \textcolor{red}{F}
### \textcolor{red}{0.50 0.50 0.00}
\end{Verbatim}

The following is an example of an unconditional generation instruction from the C24 dataset:
\begin{Verbatim}[breaklines=true,commandchars=\\\{\}]
### Instruction: Below is a description of a bulk material. Generate the space group symbol, a description of the lengths and angles of the lattice vectors and then the element type and coordinates for each atom within the lattice:
### Response: 
### \textcolor{red}{Amm2}
### \textcolor{red}{4.0 2.5 11.9}
### \textcolor{red}{90 90 90}
### \textcolor{red}{C}
### \textcolor{red}{0.19 0.00 0.67}
### \textcolor{red}{C}
### \textcolor{red}{0.29 0.00 0.98}
### \textcolor{red}{C}
### \textcolor{red}{0.30 0.00 0.23}
### \textcolor{red}{C}
### \textcolor{red}{0.31 0.00 0.55}
### \textcolor{red}{C}
### \textcolor{red}{0.00 0.00 0.32}
### \textcolor{red}{C}
### \textcolor{red}{0.00 0.00 0.90}
\end{Verbatim}

The following is an example of a crystal property prediction instruction from the MP30 dataset:
\begin{Verbatim}[breaklines=true,commandchars=\\\{\}]
### Instruction: Below is a partial description of a bulk material where the \textbf{energy above the convex hull} has been replaced with the string "[MASK]":
### The \textbf{energy above the convex hull} is [MASK].
### \textcolor{blue}{P4/mmm}
### \textcolor{blue}{4.0 4.0 10.1}
### \textcolor{blue}{90 90 90}
### \textcolor{blue}{Rb}
### \textcolor{blue}{0.00 0.00 0.50}
### \textcolor{blue}{Zn}
### \textcolor{blue}{0.00 0.50 0.16}
### \textcolor{blue}{P}
### \textcolor{blue}{0.50 0.50 0.28}
### \textcolor{blue}{P}
### \textcolor{blue}{0.00 0.00 0.00}
### Generate the \textbf{energy above the convex hull} that could replace [MASK] in the bulk material:
### Response: \textcolor{red}{0.0}
\end{Verbatim}

\section{Evaluation on Physical and Chemical Realism}
To more comprehensively evaluate the physical and chemical realism of generated crystal structures, we supplement the success rate metric with three additional evaluation criteria: atomic overlap, symmetry adherence, and energy-based feasibility check. Atomic overlap reflects the extent to which generated atoms are placed unrealistically close to one another, indicating physical invalidity of the resulting structure. Symmetry adherence measures how well the generated structure conforms to the intended space group, reflecting consistency with the specified generation condition. Energy-based feasibility check evaluates whether the generated crystal exhibits a negative formation energy, which is a necessary condition for thermodynamic stability.

\begin{table}[h]
    \centering
    \resizebox{\columnwidth}{!}{
    \begin{tabular}{c|cc|cc|cc}
        \toprule
        \multirow{2}{*}{Method} & \multicolumn{2}{c|}{Atomic Overlap $\downarrow$} & \multicolumn{2}{c|}{Symmetry Adherence $\uparrow$} & \multicolumn{2}{c}{energy-based feasibility check $\uparrow$} \\
        ~ & Mean & Std. & Mean & Std. & Mean & Std. \\
        \midrule
CrystalLLM (XYZ) & 0.0974 & 0.0100 & 0.2562 & 0.0090 & 0.8179 & 0.0205 \\
CrystalICL (XYZ) 0-Shot & 0.0840 & 0.0087 & 0.3102 & 0.0213 & 0.8411 & 0.0049 \\
CrystalICL (XYZ) 3-Shot & \textbf{0.0731} & 0.0052 & \textbf{0.3492} & 0.0188 & \textbf{0.8795} & 0.0091 \\
CrystalLLM (SGS) & 0.0868 & 0.0067 & 0.8290 & 0.0091 & 0.7753 & 0.0161 \\
CrystalICL (SGS) 0-Shot & 0.0390 & 0.0037 & 0.9600 & 0.0028 & 0.8776 & 0.0170 \\
CrystalICL (SGS) 3-Shot & \textbf{0.0290} & 0.0058 & \textbf{0.9666} & 0.0054 & \textbf{0.9151} & 0.0124 \\
        \bottomrule
    \end{tabular}
    }
    \caption{Evaluation of physical and chemical realism on MP20.}\label{extended_metrics}
\end{table}

To assess physical and chemical realism, we apply these metrics in a conditional generation task on the MP20 dataset. As shown in Table~\ref{extended_metrics}, our method consistently achieves stronger performance across all three metrics. In the few-shot SGS setting, the model generates structures with notably reduced atomic overlap and a symmetry adherence success rate exceeding 96\%. Furthermore, over 91.5\% of the generated crystals exhibit negative formation energy. These results demonstrate that the proposed SGS representation and instruction tuning strategies significantly improve the structural validity and physical plausibility of the outputs.

\section{License}
Llama-2 \cite{llama2} is distributed under the \textit{LLaMA 2 Community License Agreement} by Meta. The MEGNet model \cite{MEGNet} is released under the \textit{BSD-3-Clause license}. The pymatgen \cite{pymatgen} is released under the \textit{MIT license}. 
The Materials Project \cite{MP20} dataset, the P5 \cite{P5} dataset and the C24 \cite{C24} dataset are released under the \textit{Creative Commons Attribution 4.0 license}. 
All resources are used solely for academic research, in accordance with their licensing terms.

\end{document}